\documentclass[sigconf]{acmart}
\usepackage{amsmath}
\usepackage{amssymb}
\usepackage{adjustbox}
\usepackage{multirow}
\usepackage{subfigure}
\usepackage{balance}

\AtBeginDocument{%
  \providecommand\BibTeX{{%
    \normalfont B\kern-0.5em{\scshape i\kern-0.25em b}\kern-0.8em\TeX}}}

\copyrightyear{2021}
\acmYear{2021}
\setcopyright{acmcopyright}
\acmConference[MM '21]{Proceedings of the 29th ACM International Conference on Multimedia}{October 20--24, 2021}{Virtual Event, China}
\acmBooktitle{Proceedings of the 29th ACM International Conference on Multimedia (MM '21), October 20--24, 2021, Virtual Event, China}
\acmPrice{15.00}
\acmDOI{10.1145/3474085.3475674}
\acmISBN{978-1-4503-8651-7/21/10}

\begin{document}
\fancyhead{}

\title{Self-Supervised Regional and Temporal Auxiliary Tasks for Facial Action Unit Recognition}

\author{Jingwei Yan}
\authornotemark[1]
\email{yanjingwei@hikvision.com}
\affiliation{%
  \institution{Hikvision Research Institute}
  \city{Hangzhou}
  \country{China}
}

\author{Jingjing Wang}
\authornote{Both authors contributed equally to this research.}
\email{wangjingjing9@hikvision.com}
\affiliation{%
  \institution{Hikvision Research Institute}
  \city{Hangzhou}
  \country{China}
}

\author{Qiang Li}
\email{liqiang23@hikvision.com}
\affiliation{%
  \institution{Hikvision Research Institute}
  \city{Hangzhou}
  \country{China}
}

\author{Chunmao Wang}
\email{wangchunmao@hikvision.com}
\affiliation{%
  \institution{Hikvision Research Institute}
  \city{Hangzhou}
  \country{China}
}

\author{Shiliang Pu}
\authornote{Corresponding author.}
\email{pushiliang.hri@hikvision.com}
\affiliation{%
  \institution{Hikvision Research Institute}
  \city{Hangzhou}
  \country{China}
}


\begin{abstract}
Automatic facial action unit (AU) recognition is a challenging task due to the scarcity of manual annotations. To alleviate this problem, a large amount of efforts has been dedicated to exploiting various methods which leverage numerous unlabeled data. However, many aspects with regard to some unique properties of AUs, such as the regional and relational characteristics, are not sufficiently explored in previous works. Motivated by this, we take the AU properties into consideration and propose two auxiliary AU related tasks to bridge the gap between limited annotations and the model performance in a self-supervised manner via the unlabeled data. Specifically, to enhance the discrimination of regional features with AU relation embedding, we design a task of RoI inpainting to recover the randomly cropped AU patches. Meanwhile, a single image based optical flow estimation task is proposed to leverage the dynamic change of facial muscles and encode the motion information into the global feature representation. Based on these two self-supervised auxiliary tasks, local features, mutual relation and motion cues of AUs are better captured in the backbone network with the proposed regional and temporal based auxiliary task learning (RTATL) framework. Extensive experiments on BP4D and DISFA demonstrate the superiority of our method and new state-of-the-art performances are achieved.
\end{abstract}



\maketitle

\section{Introduction}
Facial action units describe a series of muscle movements on human faces precisely~\cite{ekman1997face}. Based on the combination of AUs, most facial expressions can be analyzed and modeled in an objective and explicit way. Nowadays, automatic AU recognition has drawn growing attention in the affective computing and computer vision communities and has been widely applied in human-computer interaction, fatigue monitoring, deception detection, etc.

The majority of existing AU recognition models are trained in a fully supervised manner which requires a correct manual annotation for each sample. However, the annotation of AUs not only needs facial action coding expertise, but also is very time-consuming, which results in limited data with reliable annotations. On the contrary, there is a large amount of easily accessible facial expression images or videos on the Internet which contain a diverse variety of AUs. In this paper, we will delve into the unlabeled data and learn discriminative feature representation for AU recognition from the aspect of self-supervised auxiliary task learning.

\begin{figure}[tbp]
	\begin{center}
		\includegraphics[width=\columnwidth]{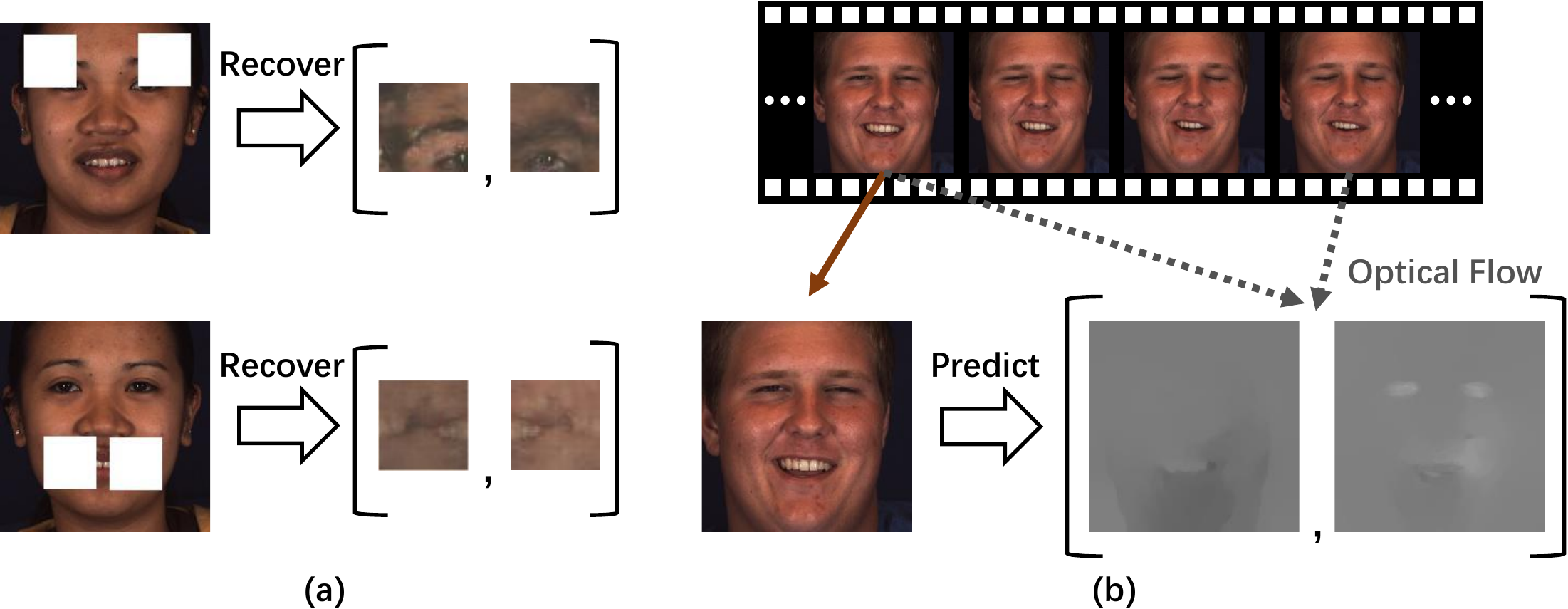}
	\end{center}
	\caption{Two proposed self-supervised auxiliary tasks. (a) RoI inpainting, which aims to recover the cropped AU RoIs via regional features embedded with AU relations. (b) Optical flow estimation, whose purpose is to explore temporal motion information of facial muscles from one static image.}
	\label{intro}
\end{figure}


Recently self-supervised learning (SSL) has shown great potential in learning discriminative features from the unlabeled data via various pretext tasks~\cite{kolesnikov2019revisiting}. For the task of AU recognition, the limited works in this promising topic mainly focused on the aspect of global feature learning, which is in accordance with some previous SSL approaches for image recognition. Li et al.~\cite{li2019self} designed the task of reconstructing one facial frame in a video sequence in order to learn features for the movement of facial muscles. Lu et al.~\cite{lu2020self} leveraged the temporal consistency to learn feature representation by a pretext task similar to examplar~\cite{dosovitskiy2014discriminative}. In both methods, SSL was conducted on the unlabeled video sequences and only global features for facial frame were explored with the proposed pretext tasks. As a result, some unique properties of AUs, which are often leveraged to obtain discriminative AU features in fully or weakly supervised methods, such as the locality of AU and relationship between AUs, are ignored in these tasks. This will lead to a nonnegligible gap between the existing self-supervised tasks and AU recognition. For the purpose of bridging this gap, we propose a regional and temporal based auxiliary task learning (RTATL) framework to incorporate regional and temporal feature learning, together with AU relation embedding simultaneously in one unified framework via two novel self-supervised tasks.

As each AU is corresponding to one or a small group of facial muscles, discriminative regional features are crucial for AU recognition. Individual convolutional layers are commonly applied on the region of interest (RoI) for different facial structures or textures. Meanwhile, the strong relationship between AUs is often leveraged to boost the recognition performance. Different from previous graph propagation methods which were based on the fixed prior AU relation graph, here a transformer~\cite{vaswani2017attention} is employed to encode the AU relation adaptively. By means of the self-attention mechanism in the transformer, relevant AU RoI features are aggregated to improve the representation capacity. However, the AU relation learned from limited labeled data may be flawed. In order to acquire accurate AU relation and enhance the representation ability of regional features through huge amounts of unlabeled data, a novel RoI inpainting (RoII) self-supervised task is designed. As shown in Figure~\ref{intro}(a), RoIs from the symmetrical parts of a random AU are cropped from the original image and then recovered based on the representative features of the copped patches. To obtain such features, other RoI features of the intact regions are leveraged to fuse together with the cropped ones via the learned AU relationship.



For some AUs like AU7 (lid tightener) and AU24 (lip presser), it is ambiguous to recognize them with the static facial image alone, especially when the intensity is weak, as the difference between low-intensity AUs and a neutral face can hardly be modeled via a single image. Therefore, it is beneficial to take the motion information of facial muscles into consideration. However, a sequence of frames is necessary to extract the temporal features conventionally and the computational costs of current multi-frame based AU recognition methods, such as 3D-CNN~\cite{tran2015learning} and CNN-LSTM~\cite{chu2016modeling}, are much heavier than image based models. To overcome the obstacle, given that the optical flow extracted from two frames naturally depicts such temporal change, we propose a self-supervised task of image based optical flow estimation (OFE), which is shown in Figure~\ref{intro}(b). The optical flow between two frames is extracted and served as the supervisory signal. Global features of the image are utilized to predict the optical flow so that the backbone network is forced to explore the muscle motions from the static image. Meanwhile, as the movements of the facial muscles only occur in certain regions caused by AU activation, the optical flow for supervision is sparse and only contains values in corresponding local facial parts. Therefore, another potential benefit of OFE is that the model will learn to focus on the important local regions automatically.

By integrating the two proposed auxiliary tasks, three crucial components for AU recognition, i.e., regional and temporal feature learning, together with AU relation encoding, are unified in one framework for the first time. Moreover, different from previous SSL methods which were conducted in two separate steps, i.e., training the feature extractor by performing the self-supervised task on unlabeled data and then training a linear classifier based on the labeled AU data, we incorporate the self-supervised tasks as auxiliary tasks which are trained with the AU recognition task simultaneously. The RTATL framework is more efficient to train as it is end-to-end trainable and can achieve much better recognition performances compared to SSL based methods.


In summary, the major contributions of this paper are as follows. (1) We propose two novel auxiliary tasks to incorporate regional, temporal and AU relation learning in a self-supervised manner. (2) Transformer is employed to encode the AU relation via self-attention mechanism in a data-adaptive fashion for the first time. (3) Based on the AU specific self-supervised auxiliary tasks, state-of-the-art performances are achieved on two benchmark databases and the principle of designing self-supervised tasks will shed light on the development of SSL based AU recognition.

\section{Related Work}
\subsection{Action Unit Recognition}
Many efforts have been dedicated to AU recognition for the past decades. For the purpose of performance improvement, discriminative regional feature learning, AU relationship embedding and temporal feature learning are fully explored in supervised methods.

In order to capture regional features for different facial structures, Zhao et al.~\cite{zhao2016deep} proposed a region layer which divides the feature maps into identical regions, then independent convolutions were applied to each region. Niu et al.~\cite{niu2019local} proposed the LP-Net to model the person-specific shape information and explore the relationship among local regions simultaneously. Corneanu et al.~\cite{corneanu2018deep} and Zhang et al.~\cite{zhang2019context} cropped several relatively large regions from the original facial image and employed individual CNNs on them. Li et al.~\cite{li2017action} defined AU centers based on the rough position relationship between AUs and landmarks, and proposed a region of interest network to crop regions around these AU centers. This landmark based region localization method was followed by~\cite{shao2018deep,li2019semantic}. 

\begin{figure*}[tbp]
	\begin{center}
		\includegraphics[width=0.85\textwidth]{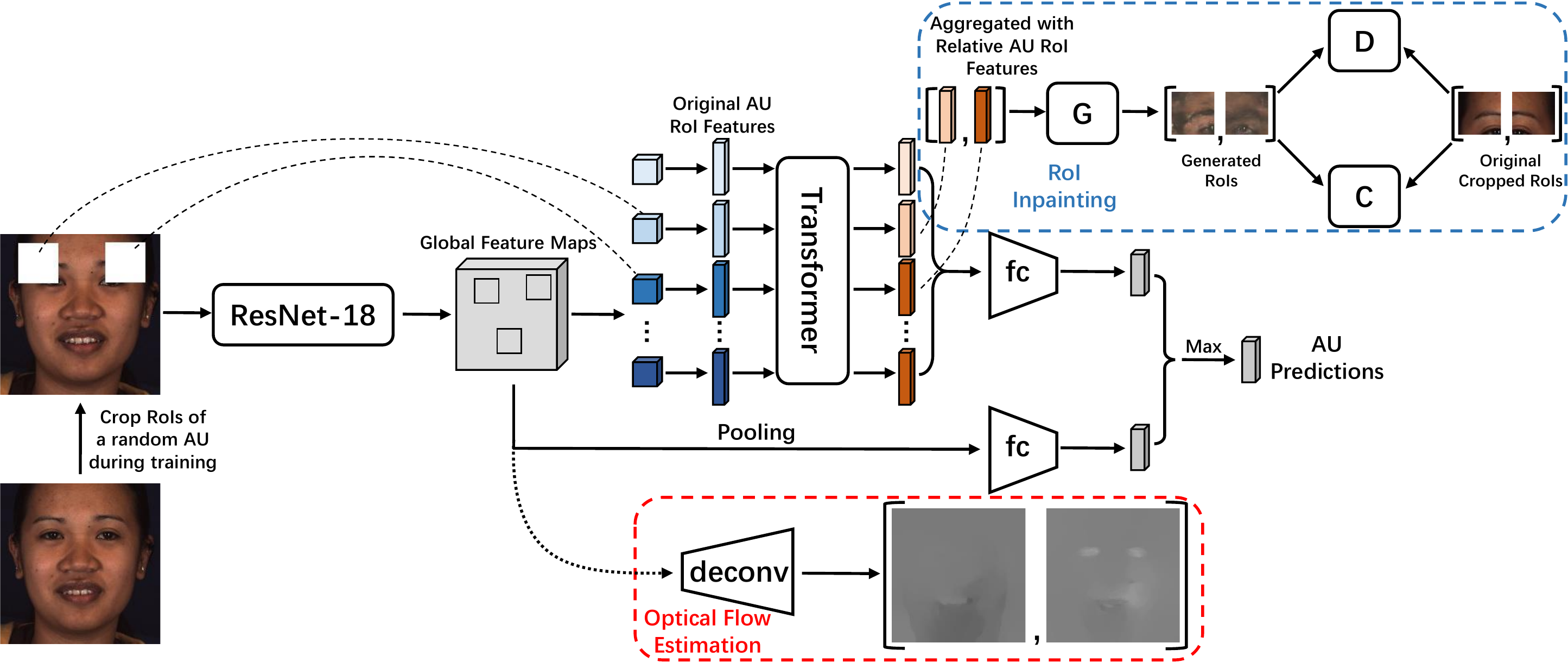}
	\end{center}
	\caption{The proposed RTATL framework. Backbone network is in the middle. The self-supervised auxiliary tasks of RoI inpainting and optical flow estimation are solved by the modules in the blue and red dashed box, respectively. During the testing phase, only the backbone network is effective and intact facial images are input to the network. Best viewed in color.}
	\label{framework}
\end{figure*}

The relationship between AUs has been exploited to boost the recognition performance. Corneanu et al.~\cite{corneanu2018deep} proposed to capture AU correlations by deep structured inference network which was applied on the AU predictions in a recurrent manner. Li et al.~\cite{li2019semantic} proposed a semantic relationship embedded representation learning (SRERL) framework to embed the relation knowledge in regional features with gated graph neural network. Graph convolutional network (GCN)~\cite{kipf2017semi} was also utilized to encode the prior AU relationship information into feature representation \cite{niu2019multi,liu2020relation,song2021uncertain,yan2021multi}. The graph propagation based methods often utilize a fixed AU relation graph to embed the relationship which is not suitable for some facial samples. In our framework, a transformer is leveraged due to the flexibility and the data driven manner and a self-supervised task is proposed to enhance the AU relation learning.

From the temporal aspect, an action unit essentially describes the motion of facial muscles. Thus it is helpful to leverage the temporal features of facial textures. To capture such dynamic change information, Chu et al.~\cite{chu2016modeling} and Jaiswal et al.~\cite{jaiswal2016deep} combined CNN and LSTM to model the AU sequence spatially and temporally. Similarly, Li et al.~\cite{li2017action} employed LSTM to fuse the temporal features of AU RoIs and improved the model performance greatly compared to the model without using temporal information. However, these methods are performed on the frame sequence and cannot be adapted to the image based models. Meanwhile, the complex network structures for temporal feature learning make the calculation inefficient during inference process. In our framework, a self-supervised auxiliary task is proposed to enhance the temporal representation capacity during training with a single static image, thus it does not increase the computational cost for AU recognition.

Some efforts were also made to integrate auxiliary tasks with AU recognition. Most of the approaches leveraged the recognition of facial expression or other related facial attributes which still required for corresponding manual annotations~\cite{yang2016multiple,chang2017fatauva,wang2017expression}. Shao et al.~\cite{shao2018deep} integrated the facial alignment task with AU recognition to learn better local features, while in our method, the two auxiliary tasks are designed to model AU characteristics from three crucial components via the unlabeled data in a self-supervised manner, which is more effective and does not need additional annotations.

\subsection{Self-Supervised Learning}
SSL methods first deduce the ground truth labels or other supervisory signals from the designed pretext task itself. Then the ground truth is leveraged to supervise the training of the feature learning model. Through the designed SSL task, generic features are learned for the downstream task. For AU recognition, current works are mainly motivated by the temporal movement of facial muscles. Wiles et al.~\cite{koepke2018self} proposed FAb-Net to generate from the source frame to target frame by low-dimensional attribute embedding. Li et al.~\cite{li2019self} also utilized the transformation between the source and target frames and furthermore proposed the twin-cycle autoencoder to obtain dedicated AU features without head motions. Inspired by the intrinsic temporal consistency in videos, Lu et al.~\cite{lu2020self} used a triplet-based ranking approach to rank the frames. These methods mainly focus on discriminative global feature learning and ignore some unique properties of AUs. Regional feature leaning and AU relation embedding are not taken into consideration. Meanwhile, these SSL tasks are performed on video samples and cannot be adopted to tremendous unlabeled facial images.

\section{Method}
In this section, the backbone network is introduced first. Adaptive AU relation embedding via the transformer is presented next. Then two proposed self-supervised tasks, i.e., RoI inpainting and optical flow estimation, are described in detail. Finally, the two tasks are further incorporated with the backbone network to form the unified end-to-end trainable RTATL framework.

\subsection{Backbone Network}

\begin{figure}[tbp]
	\begin{center}
		\includegraphics[width=0.9\columnwidth]{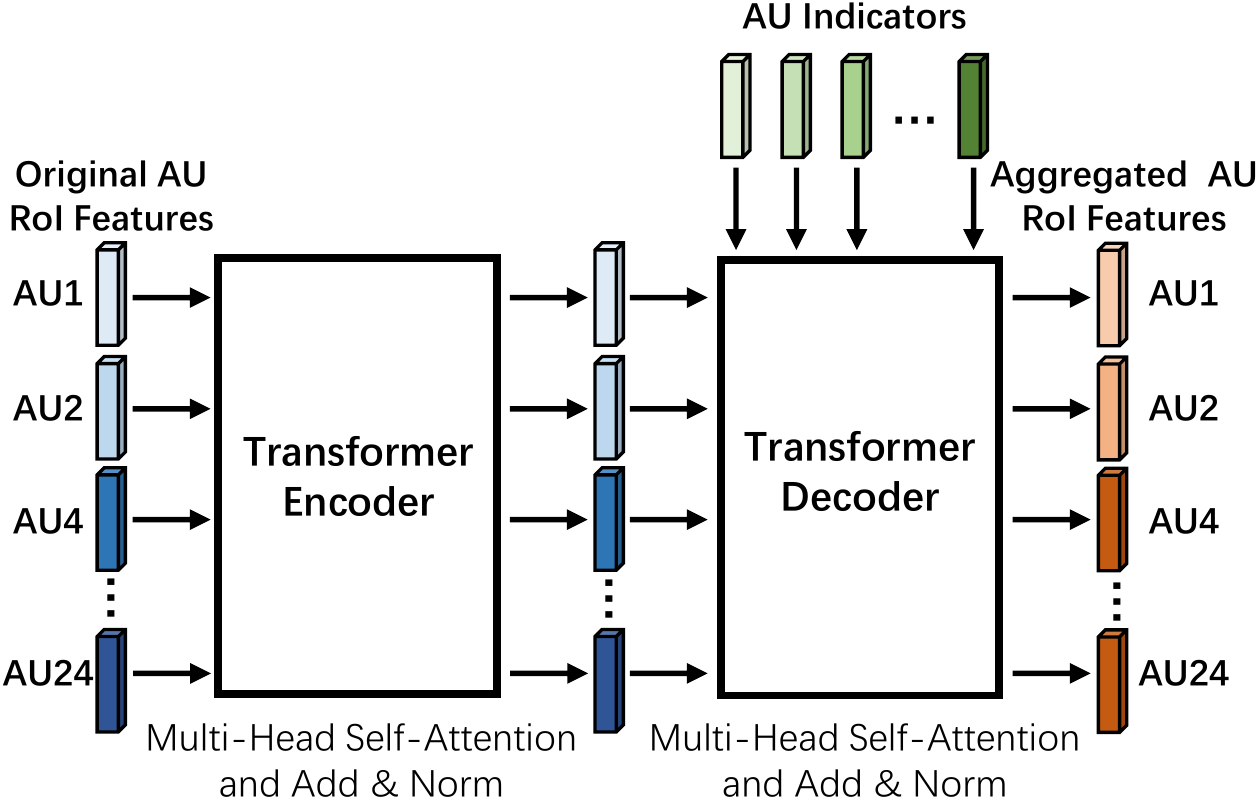}
	\end{center}
	\caption{The transformer takes vectorized RoI features as input and transforms the feature embedding with multi-head self-attention and encoder-decoder attention. AU indicators are added to the input of the decoder to obtain aggregated AU RoI features correctly.}
	\label{transformer}
\end{figure}

The backbone network is illustrated in the middle part of Figure~\ref{framework}. ResNet-18~\cite{he2016deep} is chosen as the foundation of the backbone due to the effectiveness and efficiency in feature learning. As each AU is defined on the corresponding local facial region, individual RoI feature learning module which consists of two convolutional layers for each AU is utilized to deal with different facial textures and structures. We follow the AU RoI definition proposed in~\cite{li2017action}. In order to get relatively high spatial resolution feature maps while keeping sufficient semantic information before cropping AU RoIs, the high-level feature maps are gradually upsampled and added with the low-level feature maps. Then features of each AU RoI are cropped from the fused global feature maps. Due to the symmetry of human face, each AU corresponds to two RoI features. A lightweight transformer with one encoder and one decoder is employed to model the AU relationship adaptively and encode the relation to RoI features, as shown in Figure~\ref{transformer}. We calculate the average of the two enhanced RoI features derived from the transformer for each AU and regard it as the AU RoI representation. Finally, based on local RoI features and overall global features, AU predictions made by separate fully connected layers are fused together by maximizing. 

\subsection{Transformer based Adaptive AU Relation Embedding}
The transformer was originally designed for nature language processing (NLP) tasks based on the attention mechanism~\cite{vaswani2017attention}. Compared to recurrent neutral networks, it is more parallelizable and easier to train. A typical transformer is composed of multiple encoders and decoders. For a transformer encoder, it takes the feature sequence as input and output the attention features adaptively. Feature positions in the input sequence are represented by the positional encoding. As there is no ordinal relation between AUs in the sequence, positional encoding is not added in the encoder.

As a key module in both encoders and decoders, the self-attention layer aims to explore relationship between each input feature and all others in the sequence via the self-attention mechanism. For a transformer encoder, the related AU RoI features can be attended to the corresponding one adaptively and the attention feature can be formulated as follows,
\begin{equation}
  Attention(Q,K,V)=Softmax(\frac{QK^T}{\sqrt{d}})V,
\end{equation}
where $Q$, $K$ and $V$ are linear transformations of the input feature sequence, which is formed as a matrix $X$.  In our model, feature maps obtained from RoI feature learning of each AU are vectorized first by average pooling, and then joined together to form the input feature matrix $X$. Assume there are $N$ AUs and the feature dimension of each RoI is 128, then $X \in \mathbb{R}^ {128\times N}$. After linear transformation, the feature dimension is denoted as $d$. Other modules in the encoder such as add \& norm and feed forward network are the same with the standard structure.

As shown in Figure~\ref{transformer}, the transformer decoder transforms the feature embeddings with multi-head self-attention and encoder-decoder attention. It is notable that as the decoder is permutation invariant, in order to capture the aggregated regional features for each AU correctly, $N$ positional encodings, named AU indicators, are added to the input of the decoder. Thus each output feature of the decoder can be matched with the exact corresponding AU. Similar to word embedding in NLP, AU indicators are formulated as a $d\times N$ matrix where each $d$-dimensional vector indicates one specific AU. After training the learned AU indicators should contain AU relation knowledge. We will visualize the encoded relationship in the next section.


The RoI features output from the transformer, which contain enough discriminative and representative information, are then utilized to obtain regional AU predictions and to generate the missing patches for RoI inpainting. Compared to conventional graph propagation methods, where the AU relation graph is often fixed based on prior knowledge or the statics of AU labels in database, transformer is more flexible and adaptive for each sample. The learning procedure of AU relation and feature embedding is totally data driven and does not require to build the AU relation graph manually.

\subsection{RoI Inpainting}
As AU relationship learned from the limited annotated data may not be accurate and complete, we expect to mine reasonable and comprehensive AU relation knowledge from a large amount of unlabeled data. To this end, we propose a task of RoI inpainting (RoII), which aims for recovering the cropped regions of a random AU in the original facial image. Instead of utilizing global features to generate the cropped patches as conventional image inpainting does~\cite{pathak2016context}, we leverage RoI features of the intact regions and the learned adaptive relationship to represent the cropped parts for recovery to enhance the AU relationship modeling. It is notable that some AUs share the same RoIs, such as AU12 (lip corner puller) and AU15 (lip corner depressor). As they can hardly be activated together, there is no positive correlation between them. When these shared RoIs are cropped, the RoII task can still be completed by the related RoI features.



\subsubsection{Patch Generation with AU Semantic Consistency}
As shown in Figure~\ref{framework}, the symmetrical RoIs of a random AU are first cropped from the original facial image, and then recovered in a generative and adversarial fashion~\cite{goodfellow2014generative} based on RoI features encoded with AU relationship. After the attended RoI feature $x$ of the cropped patch is obtained, we first leverage the vanilla GAN to recover the cropped RoI $p$, where $x\in \mathbb{R}^d$ and $p\in \mathbb{R}^{3 \times s\times s}$, $s$ is the height or width of $p$ in the original image and 3 denotes the RGB channels.

The network module for patch generation is shown in the blue dashed box of Figure~\ref{framework}. The cropped RoIs and the generated patches are treated as real and fake samples respectively. Discriminator $D: \mathbb{R}^{3 \times s\times s}\rightarrow \{1,0\}$ and generator $G: \mathbb{R}^d\rightarrow \mathbb{R}^{3 \times s\times s}$ are trained alternately by the adversarial loss, which is formulated as:
\begin{equation}
  \mathcal{L}_{adv}=\mathbb{E}_{p\sim P}[\log D(p)]+\mathbb{E}_{x\sim X}[\log (1-D(G(x)))],
\end{equation}
where $P=\{ p_1,p_2,...,p_n\}$ and $X=\{ x_1,x_2,...,x_n\}$ are the sets of original cropped patches and attended RoI features derived from the transformer. We train $D$ to maximize $\mathcal{L}_{adv}$ and $G$ to minimize it so that $G$ can finally generate fake samples which cannot be distinguished by $D$. Specifically, the adversarial loss for $G$ can be rewritten as:
\begin{equation}
  \mathcal{L}_{adv}^g=-\mathbb{E}_{x\sim X}[\log D(G(x))].
\end{equation}

Meanwhile, as RoII aims to recover the cropped regions, the reconstruction loss is applied to supervise $G$ during training. Here $l_1$-norm is adopted as follows,
\begin{equation}
  \mathcal{L}_{rec}=\mathbb{E}_{p\sim P, x\sim X}||p-G(x)||_1.
\end{equation}

The vanilla GAN only aims to generate fake facial patches similar to the real ones. Nevertheless, we also expect the RoI features to maintain sufficient AU discrimination besides the representative information for recovery. Thus an auxiliary AU classifier $C: \mathbb{R}^{3\times s\times s}\rightarrow \{1,0\}$ is employed for the corresponding real and fake patch pairs so that we can explicitly demand the AU semantic of the generated patch is consistent with the original one. Specifically, for unlabeled data, the pseudo label of the cropped AU output from the backbone is regarded as the semantic information. Let $\hat{y}$ denote the pseudo AU label of the cropped patch, then the objective function for the additional AU classifier $C$ is defined as:
\begin{equation}
  \mathcal{L}_C=\mathbb{E}_{p\sim P}[CE(C(p),\hat{y})],
\end{equation}
where $CE$ is the cross-entropy function. When training $G$, $C$ can provide the semantic consistency supervision for generated patches, i.e.,
\begin{equation}
  \mathcal{L}_c^g=\mathbb{E}_{x\sim X}[CE(C(G(x)),\hat{y})].
\end{equation}

Therefore, the final loss functions for $D$ and $G$ are as follows,
\begin{equation}\label{loss}
\begin{split}
  \mathcal{L}_D=&-\mathcal{L}_{adv}\\
  \mathcal{L}_G=&\lambda_1\mathcal{L}_{adv}^g+(1-\lambda_1)\mathcal{L}_{rec}+\lambda_2\mathcal{L}_c^g,
\end{split}
\end{equation}
where $\lambda_1$ and $\lambda_2$ are hyper parameters to balance the losses.

\subsection{Optical Flow Estimation}
In addition to the enhanced discriminative regional features, it is valuable to exploit the motion information of facial muscles for AU recognition. Previous SSL approaches utilized complex networks to model the transformation from the source frame to the target one to obtain the AU related feature representation. Here we propose a self-supervised task of optical flow estimation based on the static facial image to model the temporal change of facial muscles elegantly and effectively. As optical flow exactly describes the pixel movements of facial muscles between two frames, it can be served as the supervisory signals of the dynamic change of facial textures to guide the network to exploit motion information of corresponding local muscles. 

For a video sequence, we first extract the TV-L1 optical flow~\cite{zach2007duality} between two facial frames. As the frame difference between a short time step can hardly be captured while the frame after a long step is not effective for AU recognition in the current frame, the step length is set to 3 frames empirically. Meanwhile, to alleviate the effect of head motion as much as possible, we align the face in the latter frame with the transformation matrix obtained in the previous frame. The extracted optical flow is prepared before training.

As shown in the red dashed box in Figure~\ref{framework}, global feature maps output from ResNet-18 are leveraged to estimate the optical flow. In this way, the backbone network is forced to learn motion related features from the static image. The network structure for optical flow estimation is composed of 2 layers of transposed convolution which aims to gradually model the small pixel-wise displacement. Thus the objective function can be formulated as follows,
\begin{equation}
  \mathcal{L}_F=\mathbb{E}||f_g-f_p||_1,
\end{equation}
where $f_g$ and $f_p$ are the TV-L1 and predicted optical flow respectively. Similar to the reconstruction loss in the previous section, $l_1$-norm is used to measure the difference between the predicted and the TV-L1 optical flow. Another implicit benefit of OFE is that the temporal changes often occur on local facial regions, such as eyes and lip corners, so that the backbone network can also learn to focus on these important regions under this supervision.
\subsection{RTATL Framework}
In traditional SSL methods, the representation capacity of the trained model via the pretext task is evaluated by training a linear classifier with the labeled data. However, the model performances of previous SSL based AU recognition methods are not satisfactory~\cite{li2019self,lu2020self}. On the other hand, training the self-supervised task together with the downstream task end-to-end is more efficient and can obtain better results~\cite{zhai2019s4l}. Motivated by this, we incorporate the two auxiliary tasks with the supervised AU recognition task, as illustrated in Figure~\ref{framework}, and form the end-to-end trainable RTATL framework. During training, the auxiliary tasks are also performed on the labeled data. The overall loss function is formulated as:
\begin{equation}
  \mathcal{L}=\mathcal{L}_{Sup}+\mathcal{L}_D+\mathcal{L}_G+\lambda_f\mathcal{L}_F,
\end{equation}
where $\mathcal{L}_{Sup}$ is the binary cross entropy loss function which applied to the labeled data and $\lambda_f$ is the hyper parameter of loss weight for OFE. It is notable that for the image samples with cropped RoIs, we calculate $\mathcal{L}_{Sup}$ only on AUs in the intact regions. During inference phase, the intact facial image is input to the model. Only the backbone network is utilized and the modules for two auxiliary tasks are removed.

\section{Experiments}
In this section, extensive experiments are conducted on two frequently used AU recognition benchmarks to verify the effectiveness of the two proposed auxiliary tasks by comparing with other self-supervised tasks. Then we compare the integrated framework RTATL with previous state-of-the-art methods. The F1-score of single AU and the average F1-score of all AUs in each experiment are reported. Some visualization results are presented and analysed in the last part.

\begin{table*}[htbp]
	\centering
    \caption{F1-score (in \%) on BP4D database with different self-supervised tasks. The superscripts $G$ and $T$ separately denote using GCN based on fixed prior AU relation graph and employing a transformer to model AU relation adaptively. The best and the second best performances are indicated with brackets and bold font, and brackets alone, respectively.}\label{self-supervised-cmp}
	\begin{adjustbox}{max width=\textwidth}
            \begin{tabular}{c|cccccccccccc|c}
            \hline
            AU & 1 & 2 & 4 & 6 & 7 & 10 & 12 & 14 & 15 & 17 & 23 & 24 & \textbf{Avg.}\tabularnewline
            \hline
            Backbone$^{G}$ & 49.4 & 46.2 & 59.8 & 78.8 & 78.7 & 81.3 & 86.2 & 56.4 & 44.7 & 60.9 & 45.4 & 45.7 & 61.1\tabularnewline
            Backbone$^{T}$ & 49.2 & 46.3 & 58.7 & [79.7] & 78.7 & 84.2 & 87.5 & 61.5 & 51.5 & 57.8 & 47.9 & 46.8 & 62.5\tabularnewline
            \hline
            Rotation & 53.2 & 48.6 & [61.3] & 76.4 & [78.9] & 84.1 & [\textbf{88.4}] & 59.4 & 51.4 & 63.1 & [\textbf{49.0}] & 41.1 & 62.9\tabularnewline
            Exemplar & 54.2 & 45.6 & 55.9 & 77.3 & 77.8 & 83.6 & 87.6 & 63.6 & 49.8 & [64.8] & 46.7 & [52.0] & 63.2\tabularnewline
            Jigsaw & 46.7 & 36.7 & 51.4 & 75.4 & 74.1 & 82.1 & 85.3 & 59.3 & 32.7 & 59.9 & 39.6 & 40.3 & 57.0\tabularnewline
            \hline
            \textbf{OFE} & 49.0 & 46.0 & 57.1 & 77.0 & [\textbf{79.2}] & 83.0 & [88.0] & 64.1 & [52.2] & 64.7 & 50.1 & [\textbf{55.3}] & 63.8\tabularnewline
            \textbf{RoII} & [55.0] & [\textbf{53.0}] & [\textbf{62.8}] & [\textbf{79.8}] & 78.4 & [\textbf{84.6}] & 87.8 & [\textbf{64.5}] & 50.7 & 62.6 & 46.5 & 43.5 & [64.1]\tabularnewline
            \textbf{RTATL} & [\textbf{57.1}] & [49.7] & 60.5 & 77.9 & 76.1 & [84.4] & 87.2 & [64.3] & [\textbf{53.5}] & [\textbf{67.0}] & [48.9] & 48.6 & [\textbf{64.6}]\tabularnewline
            \hline
            \end{tabular}
	\end{adjustbox}
\end{table*}

\subsection{Database}
We evaluate our methods on two popular benchmark databases, i.e., BP4D \cite{zhang2013high} and DISFA \cite{mavadati2013disfa}. BP4D consists of facial expression videos collected from 41 participants, including totally 146,847 frames with valid AU labels. DISFA is composed of 27 videos. There are totally 130,815 frames and each frame is labeled with AU intensity from 0 to 5. For DISFA database, frames with intensity greater than 1 are treated as positive and the rest are negative. For all experiments, data from these two databases are treated as labeled data and we follow the protocol of subject independent 3-fold cross validation. For unlabeled data, we randomly sample 100,000 images from Emotionet~\cite{fabian2016emotionet} which is a facial expression database containing around one million facial images from the Internet.


\subsection{Implementation Details}
For each frame, we first perform face detection and alignment based on similarity transformation and obtain a $200\times 200$ RGB face image. Ordinary data augmentations such as randomly cropping and horizontally flipping are conducted to enhance the diversity. All images are resized to $192\times 192$ as the input of the network. During testing phase, only center cropping is employed.

The transformer in our model is composed of one encoder and one decoder. For all multi-head self-attention layers, the number of heads is set to 8 empirically. $d$ is set to be 128. Thus, the dimension of each AU indicator is also 128.

For RoII, we follow~\cite{li2017action} to locate the AU centers and crop a region of $48\times 48$ around each center point. Facial landmarks are detected with Dlib~\cite{king2009dlib}. $D$ and $C$ are composed of 5 layers of convolution while $G$ consists of 5 layers of transposed convolution. $\lambda_1$, $\lambda_2$ and $\lambda_f$ are empirically set as 0.1, 0.1 and 0.2, respectively. The OFE is addressed with 2 layers of transposed convolution. ResNet-18 in the backbone takes pre-trained ImageNet model weight as initialization and the rest parameters are initialized randomly. All models are trained via Adam~\cite{kingma2014adam} with the learning rate 0.0003. All experiments are implemented with PyTorch~\cite{paszke2019pytorch} and conducted on one Nvidia Tesla V100.

In the training phase, there are totally 54.62 million parameters and the computational cost is 10.75 GFLOPs. However, during inference, there are only 19.12 million parameters in the network and the computational cost is reduced to 5.57 GFLOPs since the modules working for auxiliary tasks are only effective during training, which validates the computation efficiency in model deploying.

\subsection{Comparison with Other Self-Supervised Auxiliary Tasks}
Instead of GCN, we propose to utilize a transformer to embed AU relationship adaptively. To demonstrate the superiority of the transformer, we first conduct experiments based on the backbone network with GCN and with a transformer. Here only labeled BP4D samples are used in a fully supervised manner. As shown in Table~\ref{self-supervised-cmp}, compared to GCN, the transformer improves the average F1-score by 1.4\%, which proves its advantage of adaptive AU relation learning. The backbone model with a transformer is regarded as the baseline to compare with the subsequent SSL auxiliary tasks.

Then we conduct experiments to compare the proposed self-supervised auxiliary tasks with several other successful SSL tasks in computer vision, including rotation classification~\cite{gidaris2018unsupervised}, exemplar~\cite{dosovitskiy2014discriminative} and jigsaw puzzles~\cite{noroozi2016unsupervised}. Under the same experimental setting, these ones are also served as the auxiliary tasks and jointly trained with AU recognition. Appropriate branches are added to the backbone to solve the corresponding tasks. All experiments are conducted with the same backbone except for jigsaw puzzle as it requires to divide the input image into 9 patches which is not suitable for RoI feature learning. For this task, only ResNet-18 is employed. The performances are presented in Table~\ref{self-supervised-cmp}. Since the jigsaw task divides the input image into patches which neglects the spatial relations of the joint parts, the model performance is significantly worse than others. After incorporating other SSL tasks, AU recognition results are improved to different extents. The rotation and exemplar improve the baseline by 0.4\% and 0.7\% due to better global feature representation. 

The proposed OFE task obtains an improvement of 1.3\% in average compared to the baseline and the performance on AU24 is dramatically boosted, which demonstrate the effectiveness of the motion features for certain AUs. The result is also better than examplar as the motion information provides more valuable cues for AU recognition. By enhancing the AU relation learning via RoII, the model achieves the best result among all single SSL tasks, which is 64.1\% in average and improves the baseline by 1.6\%. The performance validates the necessity of considering the unique properties of AUs when designing self-supervised tasks. Moreover, when we integrate the two tasks together and form the RTATL framework, the average F1-score is further improved by 0.5\% compared to RoII alone, as OFE and RoII aim to obtain discriminative AU features from complementary perspectives. 

\begin{figure}[tbp]
	\begin{center}
		\includegraphics[width=0.85\columnwidth]{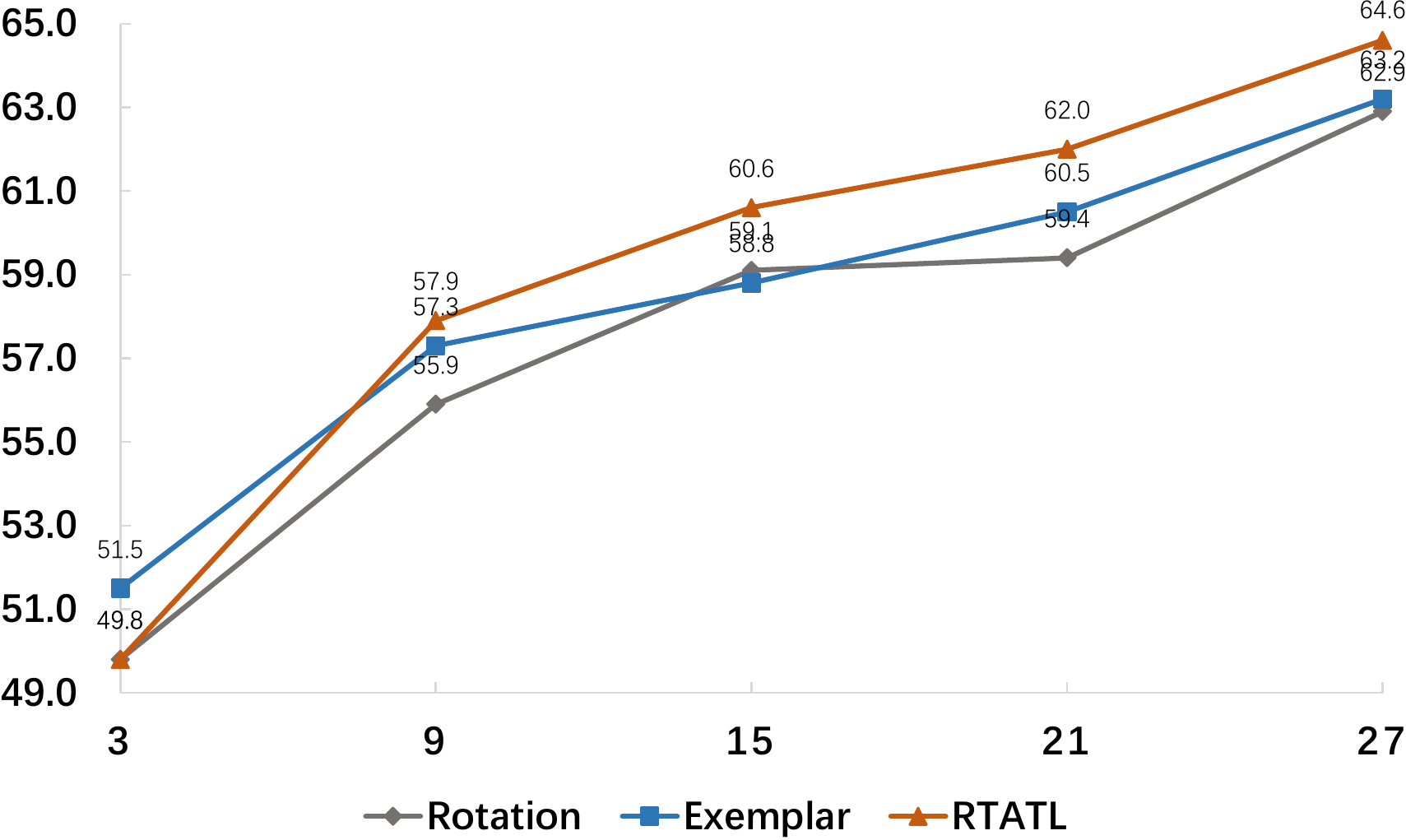}
	\end{center}
	\caption{Model performances (F1-scores, in \%) of different methods trained with an increasing number of labeled subjects, from 3 to 27 in each training set. The unlabeled data used for training is the same.}
	\label{sparse_result}
\end{figure}

Furthermore, to evaluate the model performance when only limited labeled data is accessible, we conduct experiments with a gradually increasing number of labeled training samples. Following the subject independent 3-fold cross validation protocol, instead of using all 27 subjects for training in each fold, we randomly select 3, 9, 15 and 21 subjects and discard the others in the original training set. The unlabeled and validation data are the same with previous experiments. Under the same setting, the performance comparison of three methods are shown in Figure~\ref{sparse_result}. Except for the experiment with only 3 labeled subjects, RTATL outperforms the rotation and exemplar tasks consistently. With around half the original labeled training samples, the RTATL achieves comparable result with the backbone network trained with all training samples, which demonstrates the data efficiency of our method.

\subsection{Comparison with State-of-the-art Methods}
We compare RTATL with previous state-of-the-art methods on BP4D and DISFA, including the supervised methods such as JAA~\cite{shao2018deep}, SRERL~\cite{li2019semantic}, LP-Net~\cite{niu2019local}, UGN~\cite{song2021uncertain} and the self-supervised methods such as TCAE~\cite{li2019self} and TC-Net~\cite{lu2020self}.

\begin{table}[htbp]
	\centering
    \caption{F1-score (in \%) on BP4D database.}\label{bp4d-state-of-the-art}
	\begin{adjustbox}{max width=\columnwidth}
            \begin{tabular}{c|cc|ccccc}
            \hline
            \multirow{2}{*}{AU} & \multicolumn{2}{c|}{Self-Supervised} & \multicolumn{5}{c}{Supervised}\tabularnewline
            \cline{2-8}
            ~ & TCAE & TC-Net & JAA & SRERL & LP-Net & UGN & \textbf{RTATL}\tabularnewline
            \hline
            1 & 43.1 & 42.3 & 47.2 & 46.9 & 43.4 & [54.2] & [\textbf{57.1}]\tabularnewline
            2 & 32.2 & 24.3 & 44.0 & 45.3 & 38.0 & [46.4] & [\textbf{49.7}]\tabularnewline
            4 & 44.4 & 44.1 & 54.9 & 55.6 & 54.2 & [56.8] & [\textbf{60.5}]\tabularnewline
            6 & 75.1 & 71.8 & 77.5 & [77.1] & 77.1 & 76.2 & [\textbf{77.9}]\tabularnewline
            7 & 70.5 & 67.8 & 74.6 & [\textbf{78.4}] & 76.7 & [76.7] & 76.1\tabularnewline
            10 & 80.8 & 77.6 & 84.0 & 83.5 & [83.8] & 82.4 & [\textbf{84.4}]\tabularnewline
            12 & 85.5 & 83.3 & 86.9 & [\textbf{87.6}] & 87.2 & 86.1 & [87.2]\tabularnewline
            14 & 61.8 & 61.2 & 61.9 & 63.9 & 63.3 & [\textbf{64.7}] & [64.3]\tabularnewline
            15 & 34.7 & 31.6 & 43.6 & [52.2] & 45.3 & 51.2 & [\textbf{53.5}]\tabularnewline
            17 & 58.5 & 51.6 & 60.3 & [63.9] & 60.5 & 63.1 & [\textbf{67.0}]\tabularnewline
            23 & 37.2 & 29.8 & 42.7 & 47.1 & 48.1 & [48.5] & [\textbf{48.9}]\tabularnewline
            24 & 48.7 & 38.6 & 41.9 & 53.3 & [\textbf{54.2}] & [53.6] & 48.6\tabularnewline
            \hline
            \textbf{Avg.} & 56.1 & 52.0 & 60.0 & 62.9 & 61.0 & [63.3] & [\textbf{64.6}]\tabularnewline
            \hline
            \end{tabular}
	\end{adjustbox}
\end{table}

Experimental results on BP4D are presented in Table~\ref{bp4d-state-of-the-art}. Obviously there is a performance gap between previous SSL and our method. It is because the two-stage training framework does not take advantage of the labeled AU data to train the backbone network. Meanwhile, previous SSL methods mainly focus on global feature learning while our SSL tasks are more AU specific, which incorporates regional and temporal feature learning, and AU relation encoding. Compared to the fully supervised models, RTATL fully explores the unlabeled data to capture better feature representation via the auxiliary SSL tasks, resulting in an improved performance. In comparison with other auxiliary task, such as face alignment in JAA, ours does not require additional annotations but gets a better result. It is notable that compared to the semi-supervised method MLCR~\cite{niu2019multi} which also leverages the unlabeled data, RTATL still outperforms it by 4.8\% and sets a new state-of-the-art performance.


\begin{table}[htbp]
	\centering
    \caption{F1-score (in \%) on DISFA database.}\label{disfa}
	\begin{adjustbox}{max width=\columnwidth}
            \begin{tabular}{c|cc|ccccc}
            \hline
            \multirow{2}{*}{AU} & \multicolumn{2}{c|}{Self-Supervised} & \multicolumn{5}{c}{Supervised}\tabularnewline
            \cline{2-8}
            ~ & TCAE & TC-Net & JAA & SRERL & LP-Net & UGN & \textbf{RTATL}\tabularnewline
            \hline
            1 & 15.1 & 18.7 & 43.7 & [45.7] & 29.9 & 43.3 & [\textbf{57.8}]\tabularnewline
            2 & 15.2 & 27.4 & 46.2 & 47.8 & 24.7 & [48.1] & [\textbf{52.8}]\tabularnewline
            4 & 50.5 & 35.1 & 56.0 & 59.6 & [\textbf{72.7}] & 63.4 & [70.8]\tabularnewline
            6 & 48.7 & 33.6 & 41.4 & 47.1 & 46.8 & [49.5] & [\textbf{53.2}]\tabularnewline
            9 & 23.3 & 20.7 & 44.7 & 45.6 & [49.6] & 48.2 & [\textbf{52.7}]\tabularnewline
            12 & 72.1 & 67.5 & 69.6 & [73.5] & 72.9 & 72.9 & [\textbf{74.5}]\tabularnewline
            25 & 82.1 & 68.0 & 88.3 & 84.3 & [\textbf{93.8}] & 90.8 & [91.5]\tabularnewline
            26 & 52.9 & 43.8 & 58.4 & 43.6 & [\textbf{65.0}] & [59.0] & 51.9\tabularnewline
            \hline
            \textbf{Avg.} & 45.0 & 39.4 & 56.0 & 55.9 & 56.9 & [60.0] & [\textbf{63.1}]\tabularnewline
            \hline
            \end{tabular}
	\end{adjustbox}
\end{table}

Table~\ref{disfa} presents the experimental results on DISFA. Compared to the previous state-of-the-art supervised method UGN, RTATL achieves an improvement of 3.1\% and obtains the overall performance of 63.1\%. In comparison with BP4D, DISFA contains less training samples with more severe data imbalance problem. Most of the previous methods suffer from significant performance degradation. However, in our method, by incorporating RoII and OFE in a unified framework, both local and global features in the backbone are enhanced. Therefore, RTATL maintains competitive performances on majority of the AUs consistently and achieves the new state-of-the-art performance on DISFA.

\begin{figure*}[tbp]
	\begin{center}
		\includegraphics[width=\textwidth]{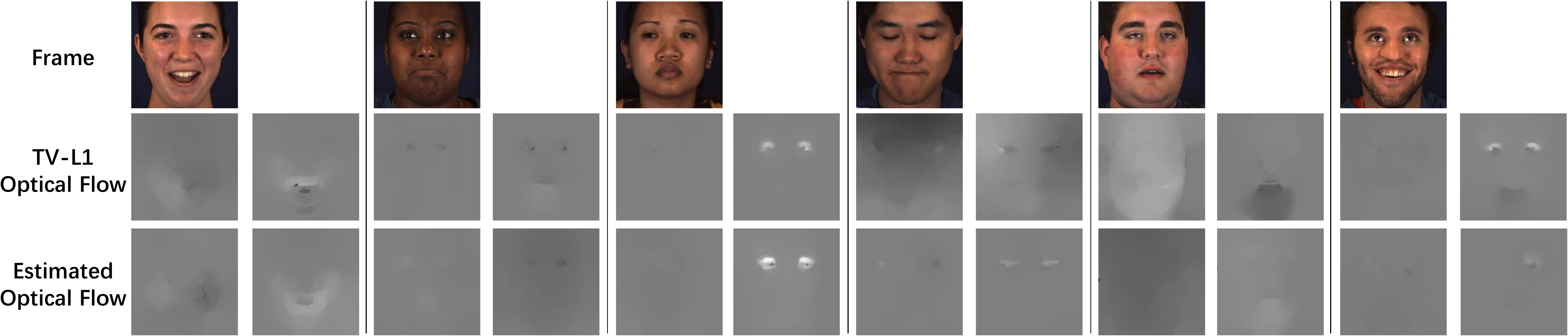}
	\end{center}
	\caption{Visualization for optical flow estimation on BP4D samples.}
	\label{flow}
\end{figure*}

\begin{figure}[tbp]
	\begin{center}
		\includegraphics[width=0.75\columnwidth]{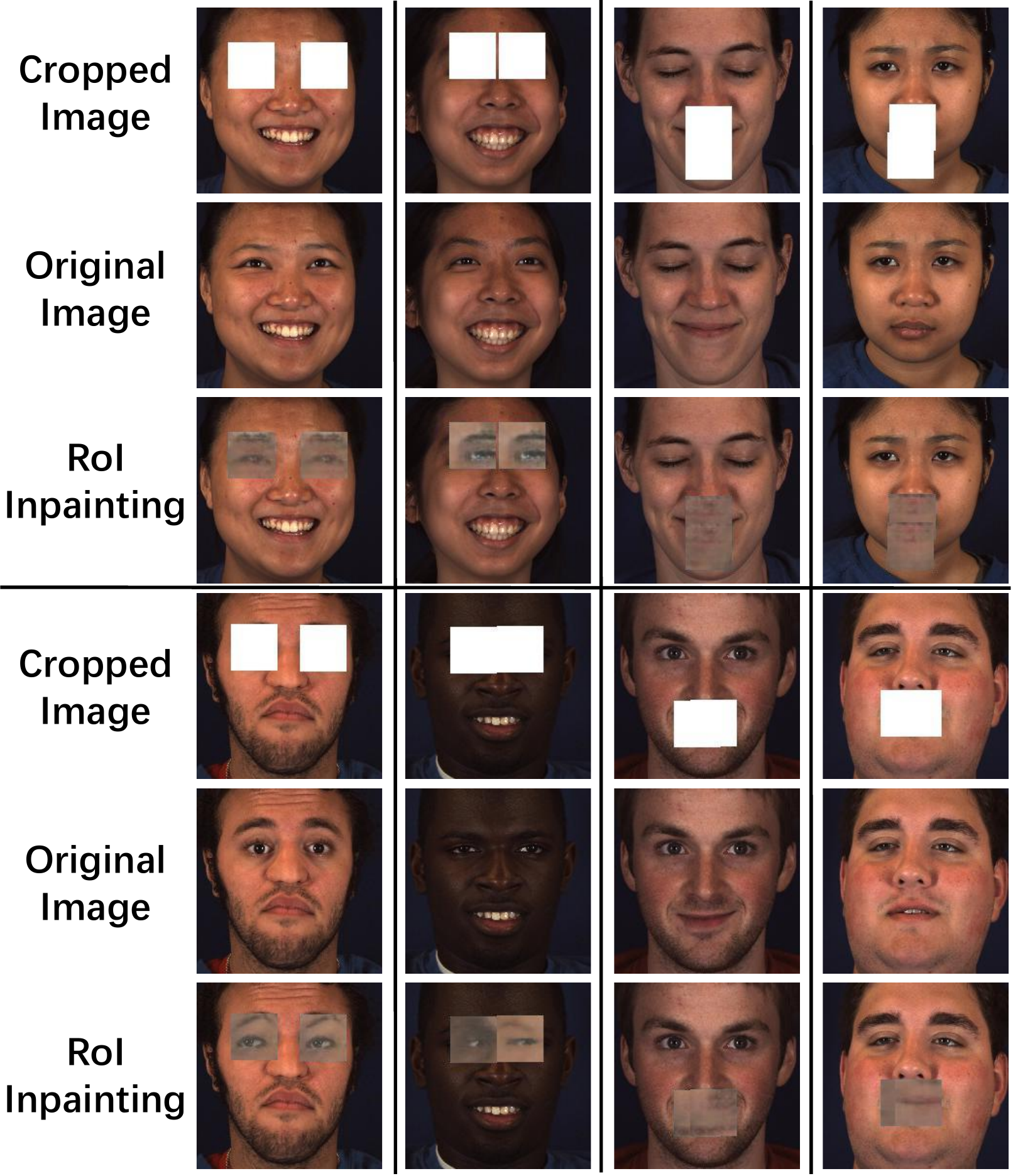}
	\end{center}
	\caption{Visualization for RoI inpainting on BP4D samples.}
	\label{inpainting}
\end{figure}

\subsection{Visualization Analysis}
The estimation of optical flow for 6 subjects in BP4D is illustrated in Figure~\ref{flow}. We use grayscale to visualize the optical flow in horizontal and vertical directions respectively, where $I_x$ is shown in left and $I_y$ is in right. TV-L1 optical flow extracted between two frames is treated as ground truth for the task of single image based OFE. As can be seen, optical flow describes the motion information of important facial regions where an AU is activated. The predictions are shown in the third row. Most of the predictions are similar to the ground truth which indicates the learned temporal features encoded in the global features are effective. Thus, through the OFE task, the backbone network can extract temporal features from single image to further enhance the discrimination of global features for AU recognition.

The visualization for RoI inpainting is shown in Figure~\ref{inpainting}. In the first row, symmetrical RoIs of a random AU are removed and replaced with white color. The original images are shown below. The third row presents the results of RoI inpainting. From the aspect of facial semantic, most of the regions around eyes and lips are recovered correctly which verifies the effectiveness of RoI features embedded with adaptive AU relation via the transformer. By the auxiliary task of RoI inpainting, the representation and discrimination of regional features are enhanced. However, as the dimension of regional feature used for recovery is only 128, and the network module for patch generation utilized is not sophisticated, some generated regions may seem poor in appearance and suffer from misalignment or mode collapse, e.g., some left and right eyes are not distinguishable. Nevertheless, the proposed RoII task aims to better explore the AU relation adaptively and embed it to the regional feature representation rather than pursue great facial appearance details in the color space. In other words, the muscle movements of the cropped region derived from other intact regions via AU relation embedding matters more in this task. From this perspective, the RoII works well to meet our demand.

\begin{figure}[tbp]
	\centering
	\subfigure[BP4D]
	{
		\begin{minipage}{0.47\columnwidth}
   			\includegraphics[width=\columnwidth]{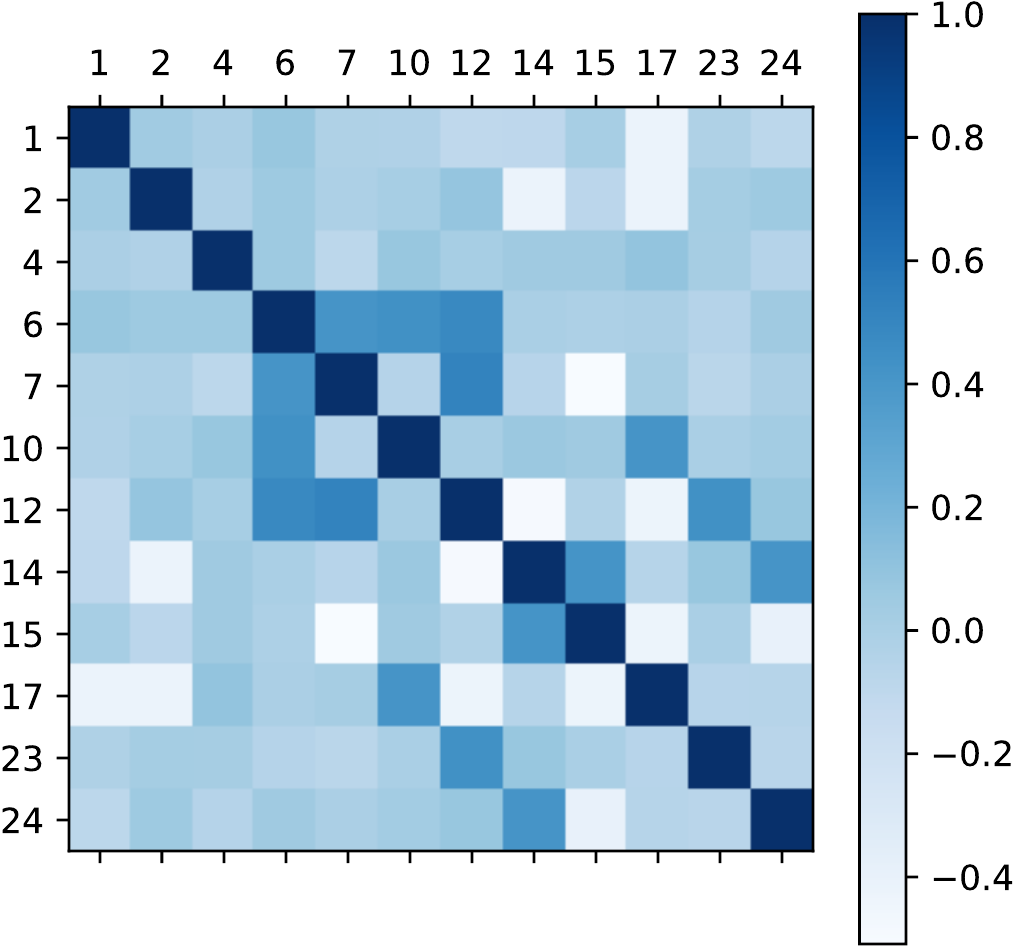}
		\end{minipage}
	}
	\subfigure[DISFA]
	{
		\begin{minipage}{0.47\columnwidth}
			\includegraphics[width=\columnwidth]{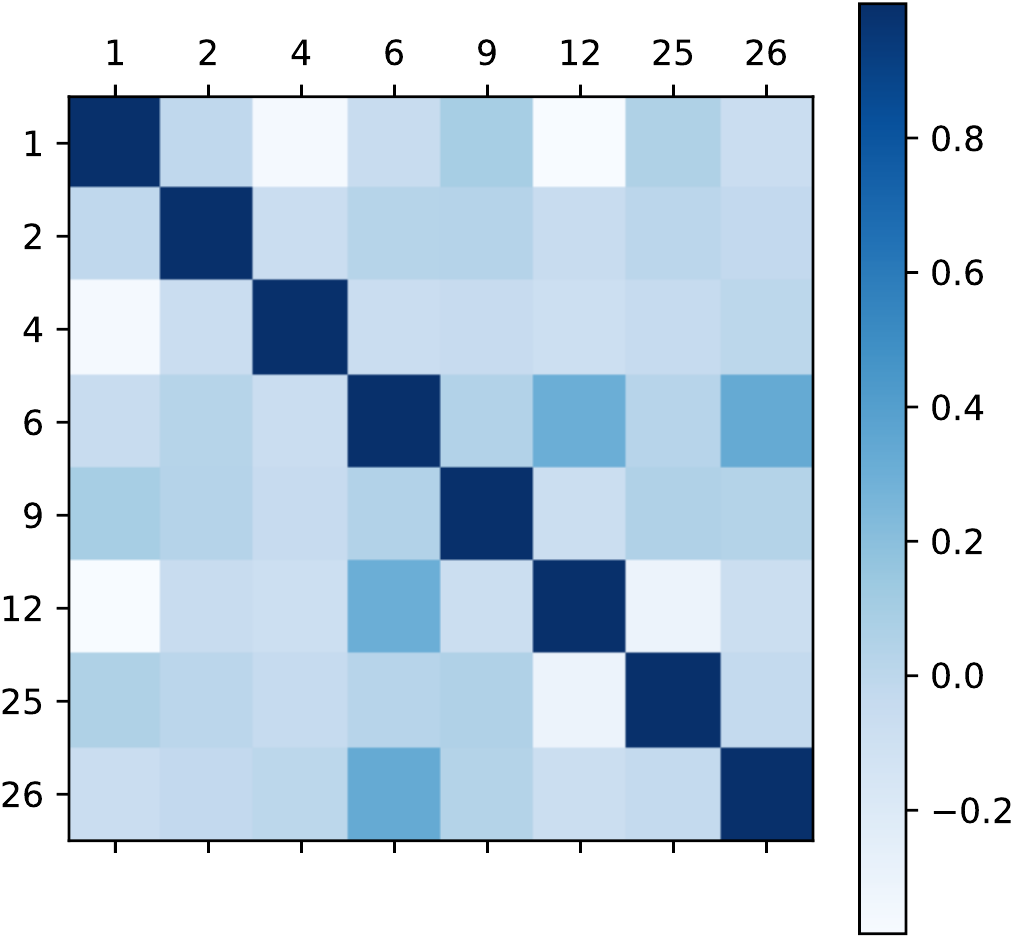}
		\end{minipage}
	}
	\caption{Cosine similarity matrices of AU indicators in the transformer decoders of RTATL for BP4D and DISFA respectively, which indicate the learned relationship knowledge between AUs. Best viewed in color.}
	\label{cos_sim}
\end{figure}

As elaborated in Section 3.2, AU indicators in the transformer decoder should contain AU relation knowledge implicitly after training. In order to visualize the learned AU relation, we calculate the cosine similarity between each AU indicator and show the similarity matrix with heatmap in Figure~\ref{cos_sim}. The relation of two AUs is closer if the blue color of the corresponding region is darker. Most of the relationships presented in the Figure~\ref{cos_sim} are consistent with the facial action coding expertise~\cite{peng2018weakly}, e.g., AU10 (upper lip raiser) and AU17 (chin raiser) are related in BP4D, and AU6 (cheek raiser and lid compressor), AU12 and AU26 (jaw drop) are closely related in DISFA. Moreover, the similarity matrix of BP4D also reveals some unique relations in the database, e.g., there exists close relationship between AU6, AU7 (lid tightener), AU10 and AU12. It is because AU6 and AU7 are very similar, and the annotators of BP4D tend to annotate AU10 as positive when the subject is laughing which is reasonable. Therefore, the transformer in our model successfully captures the AU relation adaptively in a data-driven manner.

\section{Conclusion}
In this paper, motivated by the unique properties of AUs, we take the regional and temporal feature learning, together with AU relation embedding into consideration, and design two novel auxiliary tasks in a self-supervised manner to leverage large amounts of unlabeled data to improve the performance of AU recognition. Meanwhile, a transformer is utilized to embed the AU relation adaptively for the first time. Furthermore, we incorporate the two tasks with AU recognition and propose the joint framework RTATL, which is end-to-end trainable and can capture more discriminative features from labeled and unlabeled data. Extensive experimental results demonstrate the superiority of our method and new state-of-the-art performances on BP4D and DISFA have been achieved.

\bibliographystyle{ACM-Reference-Format}
\balance
\bibliography{sample-base}


\end{document}